# When Less Is More: A Controlled Benchmark of Lightweight CNNs for Satellite Land-Cover Segmentation on DeepGlobe

**Atiq Ur Rehman[1], Joseph Michael Donovan[1]**

**[1]University of South Dakota, Vermillion, USA**

## Abstract

High-resolution satellite imagery is the backbone of good land-cover classification, and without that, environmental monitoring, urban planning, and sustainable resource management all fall short. Deep learning architectures perform well in semantic segmentation, but the efficiency-accuracy trade-off across classical convolutional encoders is not well quantified under controlled, reproducible conditions. This study compares five architectures VGG16, MobileNetV2, InceptionV3, AlexNet, and CNN on the DeepGlobe Land Cover Classification dataset using three progressively optimized iterations to isolate regularisation, transfer learning, and architectural depth. To ensure performance differentials reflect architectural properties, all experiments used identical preprocessing, hyperparameter, and training protocols without data augmentation or class-imbalance correction. At 24.98 MB, MobileNetV2_v1 had the highest overall accuracy (0.7906) and mean Intersection over Union (0.4625), outperforming deeper alternatives like InceptionV3_v2 (125.17 MB, accuracy 0.7610) and VGG16_v2 (71.13 MB, accuracy 0.7653). Class-wise analysis showed strength in urban, agricultural, and water categories, but rangeland–barren confusion showed that architectural optimization alone cannot optimize spectrally similar minority classes. Strong spatial generalization and crisp boundary delineation were confirmed on held-out test imagery, validating operational applicability. These results show that lightweight, transfer-learned models can match or outperform deeper models in resource-constrained remote-sensing environments, enabling scalable land-cover mapping.

Corresponding Author: Atiq (atiqur.rehman@usd.edu/atiq.ur.rehman.shafeeq@gmail.com). Affiliation: Department of Economics & Decision Sciences, University of South Dakota, Vermillion, SD, USA.
https://github.com/AtiqML/DeepGlobe

**Keywords:** Semantic segmentation, DeepGlobe, MobileNetV2, transfer learning, remote sensing

# Introduction

Land-use and land-cover (LULC) classification captures how human activity shapes the environment, and getting remote sensing images classified correctly is fundamental to any serious land-use and land-cover change research (Jin et al., 2019). When land-cover classification is done precisely, we gain much clearer insight into environmental change, urbanisation, agricultural output, and resource management especially now that high-resolution satellite imagery lets us observe Earth's surface in considerable detail (Ullah et al., 2025). That said, traditional statistical and pixel-based methods such as Maximum Likelihood Estimation and Support Vector Machines (SVMs) often struggle to handle the complex spatial-spectral relationships found in modern remote-sensing data (Mohanty et al., 2025; Das & Goswami, 2025). Conventional machine learning algorithms also tend to generalise poorly across intricate spatial-spectral patterns, and the problem only gets worse as intra-class diversity grows and inter-class similarity increases because of spectral confusion in high-resolution images (Bhatt et al., 2021).

Recent progress in deep learning and Convolutional Neural Networks (CNNs) has made it possible to automate hierarchical feature extraction and achieve better semantic segmentation (Acuña-Alonso et al., 2024; Rangel et al., 2024). Researchers have tested U-Net models with ResNet34, InceptionV3, and VGG16 encoders for high-resolution land-cover categorisation, with ResNet34 reaching a validation accuracy of 0.85 and InceptionV3 showing stronger generalisability (Kumar et al., 2024). InceptionV3 hit 94.36% accuracy on the UC Merced dataset through a multi-branch convolutional design combined with dropout and early stopping (Alem & Kumar, 2022), while a U-Net with a pre-trained VGG16 encoder averaged an IoU of 59.23% and 69.27% on the DeepGlobe dataset (Dey et al., 2024). More recently, IRUNet built on InceptionResNetV2 and U-Net for Sentinel-2 imagery achieved 98.21% accuracy and an 88.96% Dice score (Jagannathan, Vadivel, & Divya, 2025). Transfer learning and adaptive optimisation have also helped improve how well features generalise and how stable convergence becomes (Fayaz et al., 2024; Marzougui et al., 2025). The catch is that these deeper transfer-learning models tend to overfit and demand heavy computation, which makes them hard to deploy in operational settings with limited resources (Bhatt et al., 2021; Qin et al., 2022).

Lighter deep learning architectures have started to change remote sensing analysis by delivering accurate classification without the same computational burden. MobileNetV2, for instance, scored an F-beta of 0.9275 on the Planet Amazon dataset beating Vision Transformer, ResNet-50, DenseNet-121, and InceptionV3 while staying computationally efficient (Barman & Susan, 2024). Comparative work shows MobileNetV2 reaches 93.58% accuracy against EfficientNet's 94.36%, yet it is markedly more efficient for mobile and embedded use (Thapliyal et al., 2024). Hybrid ensembles combining MobileNetV2 and EfficientNetB0 have pushed land-use and land-cover classification accuracy even further (Pekala et al., 2025), and MobileNetV2-U-Net has proven useful for picking out urban open areas and green spaces (Dabra & Kumar, 2023). Self-attention-based variants of MobileNetV2 have sharpened feature localisation (Anjali et al., 2024), and U-Net with selective feature extraction using lightweight backbones has shown solid results for land cover classification (Ramos & Sappa, 2025). Taking together, these findings suggest that well-optimised lightweight architectures can provide scalable, efficient classification even when computational resources are tight.

Despite these advances, current research often lacks a uniform experimental framework for comparing models under identical conditions (Qin et al., 2022). Existing comparative research in remote sensing have focused on modern encoder-decoder pipelines such as U-Net and DeepLabV3+, or on hyperspectral patch classification, leaving the efficiency-performance frontier of classical CNN encoders, including basic CNN, AlexNet, VGG16, and InceptionV3 when adapted with minimal decoder heads for semantic segmentation on the DeepGlobe dataset, largely underexplored. Furthermore, the effect of progressive regularisation strategies, including transfer learning, partial fine-tuning, and adaptive dropout, across these architectures under controlled training conditions has not been systematically quantified.

This study addresses this gap by establishing a controlled, reproducible benchmark that prioritises deployability over absolute accuracy. Five architectures, CNN, AlexNet, InceptionV3, MobileNetV2, and VGG16, are systematically compared under a unified experimental framework on the DeepGlobe Land Cover Classification dataset (Demir et al., 2018). Each architecture is evaluated through three progressively optimised versions to isolate the marginal gain of regularisation and fine-tuning depth. Performance is assessed using Overall Accuracy, mean IoU, Precision, Recall, and F1-score to identify the optimal efficiency-accuracy frontier for resource-constrained land cover mapping. The remainder of this paper is organized as follows: the dataset and preprocessing pipeline are described next, followed by the model architecture, training strategy, and evaluation metrics. The results and analysis are then presented, before the paper concludes with limitations and future directions.

# Methodology

## Dataset

We chose the DeepGlobe Land Cover Classification dataset as our main data source because it offers high-resolution satellite imagery, annotations across multiple land-cover classes, and has been widely used to benchmark semantic segmentation models (Demir et al., 2018). The dataset contains 803 training images, 171 validation images, and 172 test images. Each RGB satellite image measures 2448 × 2448 pixels at 50 cm resolution and is paired with a pixel-level annotation mask. Seven land-cover classes are encoded via RGB values defined in a class dictionary file: urban, agricultural, rangeland, forest, water, barren, and unknown. The dataset supports large-scale, diverse terrain analysis and enables systematic comparison of architectures for efficiency and generalisability. CNNs pick up hierarchical features directly from the data on their own, which sharpens segmentation accuracy and makes it straightforward to validate results against standard metrics and compare models empirically (Minaee, et al., 2022). The data is publicly available and contains no personally identifiable information.

We ran all experiments on Google Colab Pro, using NVIDIA T4 GPUs for acceleration. Our software setup included Python 3.10, TensorFlow 2.x with Keras, plus OpenCV, NumPy, scikit-learn, Matplotlib, and Seaborn for processing, analysis, and visualisation. The DeepGlobe dataset was acquired via the Kaggle API.

## Data Preprocessing

Data preprocessing followed standard practice: inputs were normalised, RGB masks were converted to categorical indices, and all images were brought to uniform dimensions (Dehbozorgi, Ryabchykov, & Bocklitz, 2024). Both satellite images and masks were resized to 256 × 256 pixels bilinear interpolation for the images, nearest-neighbour for the masks so the labels would not get distorted. Pixel values in the RGB images were scaled to [0, 1] by dividing by 255. RGB-encoded mask pixels were converted to categorical class indices using the dataset's class dictionary. We then split the data into training, validation, and test sets. The validation set guided parameter optimisation and hyperparameter tuning, while the test set was held back entirely to measure final accuracy on data the model had never seen. Keeping these sets, separate is what lets us compare different architectures fairly and ensures the evaluation is both honest and reproducible.

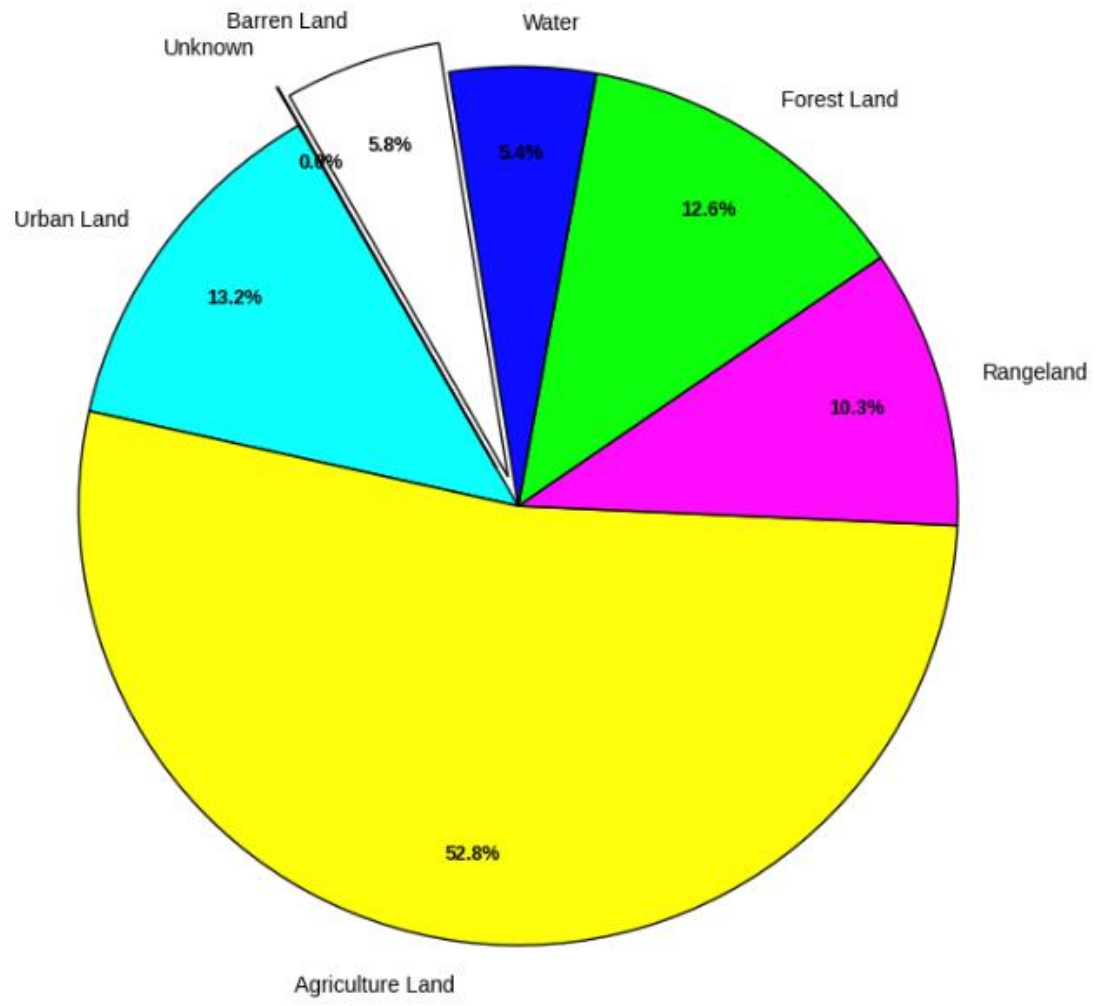

**Fig.1** Percentage breakdown of land-cover classes within the DeepGlobe dataset

The dataset exhibits significant class imbalance at the pixel level: agriculture dominates (~52.8% of labeled pixels), while urban, forest, and rangeland account for 13.2%, 12.6%, and 10.3%, respectively. Water, barren, and unknown classes are minority categories (Fig. 1). Class imbalance was deliberately left uncorrected via reweighting, focal loss, or resampling to ensure the study evaluated raw architectural discrimination capabilities under natural dataset distributions. Artificially forcing uniform class distributions during model training can severely compromise calibration metrics and distort authentic baseline boundaries (Tian et al., 2020). Furthermore, no data augmentation such as rotation, flipping, or color jitter was applied beyond baseline resizing and normalization. This restriction is supported by recent findings indicating that generic visual transformations can "substantially alter scale or spatial structure," which in turn "can introduce semantically inconsistent representations and degrade model performance" (Habrat & Młynarczuk, 2026). By omitting these artificial modifications, all recorded performance variances strictly reflect intrinsic architectural and regularization effects rather than external, augmentation-induced variability.

# Model Architectures

Five convolutional architectures were developed and compared: a custom CNN, AlexNet, InceptionV3, MobileNetV2, and VGG16. Each architecture was examined through three progressively optimised iterations: baseline (v1), enhanced (v2), and optimised (v3) to isolate the effects of regularisation, architectural depth, and transfer learning on segmentation accuracy. The baseline configurations employed shallow custom networks or frozen pretrained encoders. Enhanced versions incorporated structural modifications such as skip connections, bottleneck layers, and partial fine-tuning. Optimised versions integrated advanced regularisation, dynamic learning-rate scheduling, and progressive dropout to improve robustness and generalizability Previous work has made a strong case for structured experimentation across model variants as a way to ground evaluation in remote-sensing applications, especially where accuracy and scalability both matter (Thapliyalet al., 2024; Fayaz et al., 2024). This study sought to produce a segmentation framework for land-cover classification that was lightweight enough to be practical yet robust enough to generalise, pursued primarily through successive iterations of model refinement.

To ensure a fair comparison, each encoder was paired with a lightweight decoder progressively restoring spatial resolution to 256 × 256 pixels and terminating in a 1 × 1 convolution with SoftMax activation for seven-class prediction. Decoder complexity ranged from simple upsampling (CNN_v1, CNN_v2) to multi-layer transposed convolution stacks with batch normalisation and ReLU (InceptionV3, VGG16, MobileNetV2), scaled proportionally to encoder complexity (Table 1).

**Table 1** Model configurations and parameter count

| **Architecture** | **Version** | **Key Modification** | **Total Parameters** |
|---|---|---|---|
| **CNN** | v1 | 3 conv blocks, 1×1 + upsampling decoder | 2.96 MB |
| **CNN** | v2 | Encoder–decoder with skip connections | 62.80 MB |
| **CNN** | v3 | + SpatialDropout2D, AdamW, cosine decay | 26.45 MB |
| **AlexNet** | v1 | 5 conv layers, 512-filter bottleneck | 24.81 MB |
| **AlexNet** | v2 | + Residual skip connections, LeakyReLU | 24.81 MB |
| **AlexNet** | v3 | + Progressive dropout (0.1→0.3) | 24.81 MB |
| **InceptionV3** | v1 | Frozen ImageNet encoder | 125.10 MB |
| **InceptionV3** | v2 | Partial fine-tuning (first 150 layers frozen) | 125.17 MB |
| **InceptionV3** | v3 | + Hierarchical dropout, 1024-filter bottleneck | 179.19 MB |
| **MobileNetV2** | v1 | Frozen encoder, skip connections | 24.98 MB |
| **MobileNetV2** | v2 | Extended decoder, 512-filter bottleneck | 21.92 MB |
| **MobileNetV2** | v3 | + Separable convolutions, L2 regularisation | 18.24 MB |
| **VGG16** | v1 | Frozen ImageNet encoder | 71.13 MB |
| **VGG16** | v2 | Partial fine-tuning (blocks 4–5 unfrozen) | 71.13 MB |

| **VGG16** | v3 | + Progressive dropout (0.1→0.5) | 74.68 MB |
|---|---|---|---|

Transfer learning models (InceptionV3, MobileNetV2, and VGG16) were initialised with ImageNet-pretrained weights; the custom CNN and AlexNet were trained from random initialisation.

# Training and Validation Strategy

All experiments were conducted under controlled conditions with identical preprocessing and standardised hyperparameters across every model. A fixed random seed (42) was used for the train/validation/test split to ensure consistent data partitioning; however, seeds were not fixed for batch shuffling or model weight initialisation, so exact numerical reproduction of the reported metrics may vary slightly between runs. Training was performed for a maximum of 50 epochs with sparse categorical cross-entropy loss, a batch size of 8, and the Adam optimiser at an initial learning rate of $1 \times 10^{-4}$. To prevent overfitting and guarantee stable convergence, early stopping with a patience of 10 epochs (monitored on validation loss) was utilised alongside model checkpointing to preserve optimal weights (Rajabi et al., 2025). Batch normalisation was applied to stabilise activations, while adaptive learning rate scheduling (ReduceLROnPlateau) refined optimisation when validation loss plateaued. Dropout and L2 regularisation were incorporated in optimised configurations to suppress overfitting. Minor deviations were introduced for specific variants: CNN_v2 used a reduced learning rate of $5 \times 10^{-5}$; CNN_v3 used the AdamW optimiser with a cosine-decay schedule (initial rate $5 \times 10^{-4}$, weight decay $1 \times 10^{-4}$); and InceptionV3_v3 used a shorter early-stopping patience of 8 epochs. This systematic and reproducible training process ensured that model improvements reflected architectural robustness rather than training-related fluctuations.

**Table 2** Overview of training techniques and optimisation strategies implemented across model architectures

| Technique | Description | Applied In |
|---|---|---|
| **Dropout** | Randomly masks neurons during forward passes to curb overfitting | CNN_v3, AlexNet_v3, VGG16_v3 |
| | (Omar & El-Hafeez, 2024). | |
| **Batch Normalisation** | Recentres and rescales layer inputs to speed up training and stabilise gradients (Sagar et al., 2025). | All architectures |
| **Freezing Layers** | Locks pretrained encoder weights to retain generic low-level representations (Dobrzycki et al., 2025). | InceptionV3_v1, VGG16_v1, MobileNetV2_v1 |
| **Partial Fine-Tuning** | Thaws deeper backbone layers for task-specific feature recalibration (Benamara et al., 2025). | VGG16_v2, InceptionV3_v2 |
| **Cosine Decay** | Anneals the learning rate along a cosine curve for steadier convergence (Airen & Agrawal, 2025). | CNN_v3 |
| **ReduceLROnPlateau** | Triggers stepwise learning rate reduction upon validation stagnation (R et al., 2024). | All architectures |
| **Adam / AdamW Optimiser** | Leverages first- and second-moment estimates with decoupled weight decay (Liu et al., 2025). | CNN_v3, InceptionV3_v3 |
| **Early Stopping** | Terminates training when validation | All architectures |
| | metrics cease improving (Rajabi et al., 2025). | |
| **Model Checkpointing** | Preserves the top-performing weights across epochs for downstream reuse (Huang et al., 2025). | CNN_v3, InceptionV3_v3 |
| **SpatialDropout2D** | Nullifies complete feature maps rather than scattered units for stronger regularisation (Ninama & Raikwal, 2025). | CNN_v3 |
| **L2 Regularisation** | Adds a squared-norm penalty to the loss surface to discourage extreme | VGG16_v2, MobileNetV2_v3 |

weights (Koishiyeva et al., 2025).

## Evaluation Metrics

Segmentation performance was evaluated using Overall Accuracy, Precision, Recall, F1-score, and mean Intersection over Union (mIoU). Overall Accuracy captures the share of pixels classified correctly. Precision reflects how complete the predicted classes are, whereas Recall indicates what fraction of actual positive instances were picked up. The F1-score brings Precision and Recall together through their harmonic mean, which proves especially useful when dealing with imbalanced classes (Cho, 2024; Naidu, Zuva, & Sibanda, 2023). Mean IoU, meanwhile, gauges the spatial overlap between predicted and ground-truth masks, making it a dependable choice for multi-class segmentation tasks. Taken together, these metrics offer clear, reproducible benchmarks for refining models, improving how well they generalise, and pinpointing the architecture that performs best.

# Results and Analysis

## Comparative Performance Overview

Table 3 summarises the best-performing version for each architecture. MobileNetV2_v1 achieved the highest overall accuracy (0.7906) and mean IoU (0.4625). CNN_v3 attained the second-highest accuracy (0.7899) and mean IoU (0.4480). VGG16_v2 recorded the highest mean IoU among VGG16 variants (0.4244), while VGG16_v3 achieved marginally higher accuracy (0.7690) but lower mean IoU (0.4214). AlexNet_v2 and InceptionV3_v2 emerged as the top configurations for their respective architectures.

**Table 3** Summary of best-performing configurations per architecture

| **Architecture** | **Best Version** | **Overall Accuracy** | **Mean IoU** | **Precision (macro)** | **Recall (macro)** | **F1-score (macro)** | **Model Size** |
|---|---|---|---|---|---|---|---|
| **CNN** | v3 | 0.7899 | 0.448 | 0.6134 | 0.5475 | 0.5533 | 26.45 MB |
| **AlexNet** | v2 | 0.7649 | 0.4064 | 0.5824 | 0.4937 | 0.5183 | 24.81 MB |
| **InceptionV3** | v2 | 0.761 | 0.4118 | 0.5885 | 0.4996 | 0.532 | 125.17 MB |
| **MobileNetV2** | v1 | 0.7906 | 0.4625 | 0.6069 | 0.5641 | 0.5786 | 24.98 MB |
| **VGG16** | v2 | 0.7653 | 0.4244 | 0.6357 | 0.4996 | 0.5411 | 71.13 MB |

# CNN Results

Table 4 reports the quantitative performance of the three CNN iterations. CNN_v3 achieved the highest overall accuracy (0.7899) and mean IoU (0.4480), followed by CNN_v1 (accuracy 0.7669, mean IoU 0.4127). CNN_v2 recorded the lowest metrics across all categories.

**Table 4** Performance metrics for CNN variants

| **Model** | **Overall Accuracy** | **Mean IoU** | **Precision (macro)** | **Recall (macro)** | **F1-score (macro)** |
|---|---|---|---|---|---|
| **CNN_v1** | 0.7669 | 0.4127 | 0.5851 | 0.4975 | 0.517 |
| **CNN_v2** | 0.6019 | 0.1862 | 0.4516 | 0.2561 | 0.2615 |
| **CNN_v3** | 0.7899 | 0.448 | 0.6134 | 0.5475 | 0.5533 |

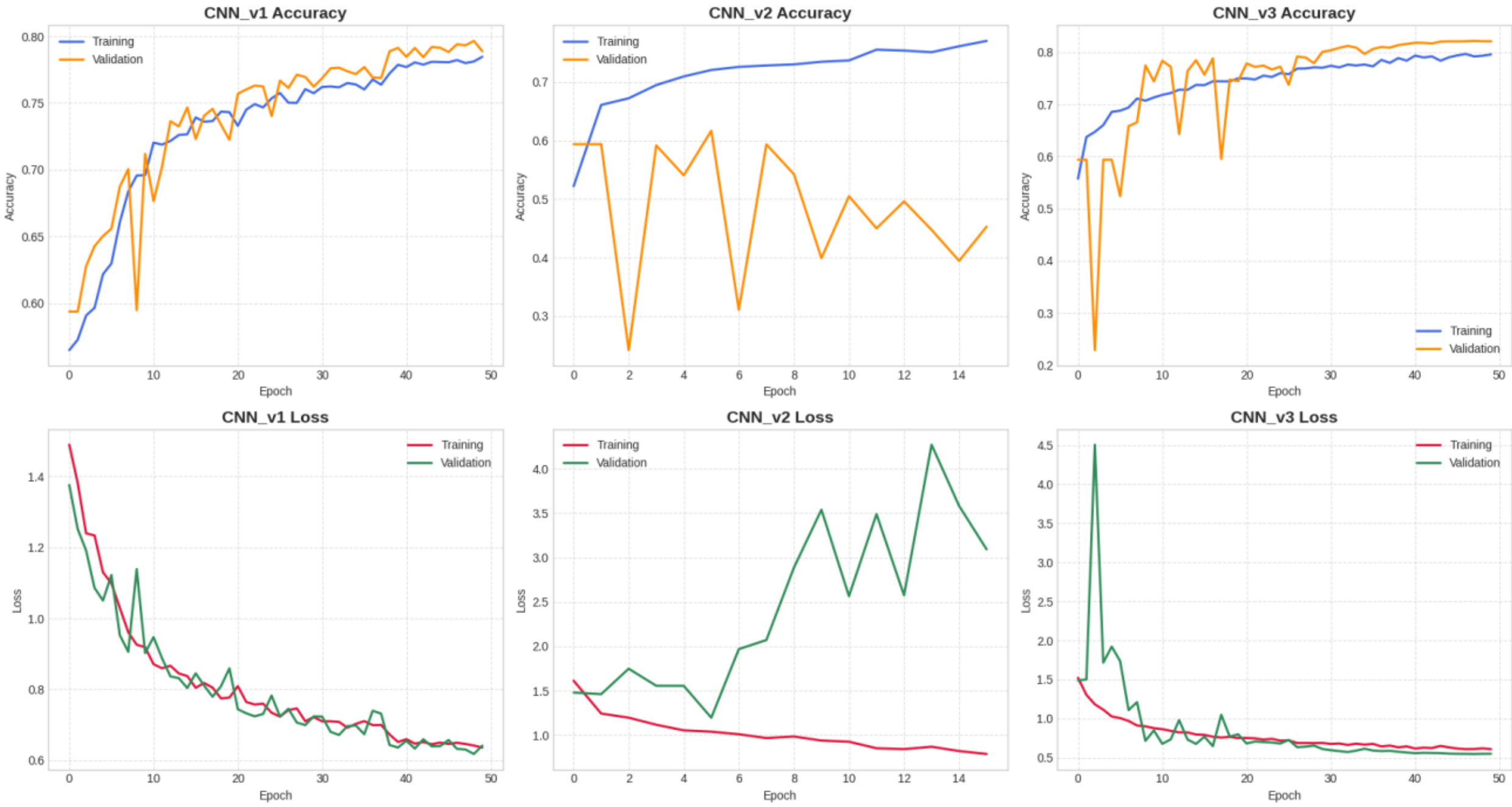

**Fig. 2** Training and validation accuracy and loss curves for the three CNN configurations (v1, v2, v3) over 50 epochs

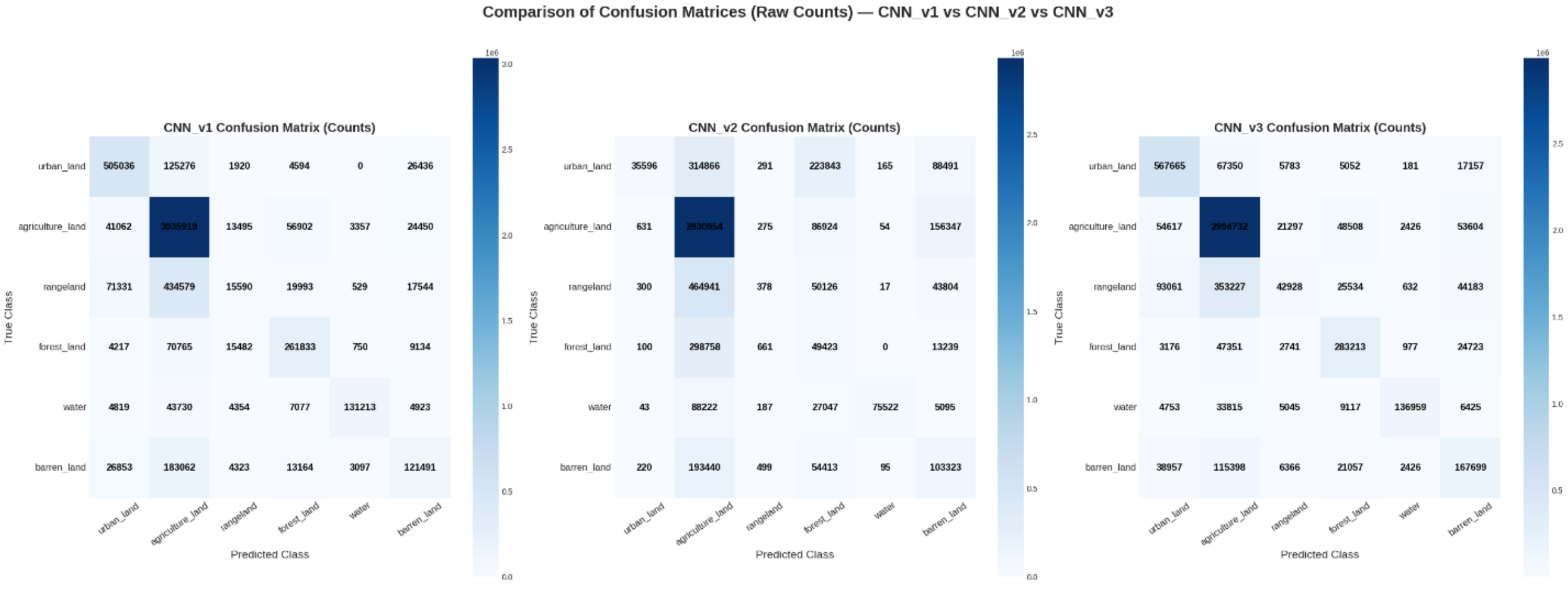

**Fig. 3** Normalised confusion matrices for CNN_v1, CNN_v2, and CNN_v3 on the DeepGlobe test set

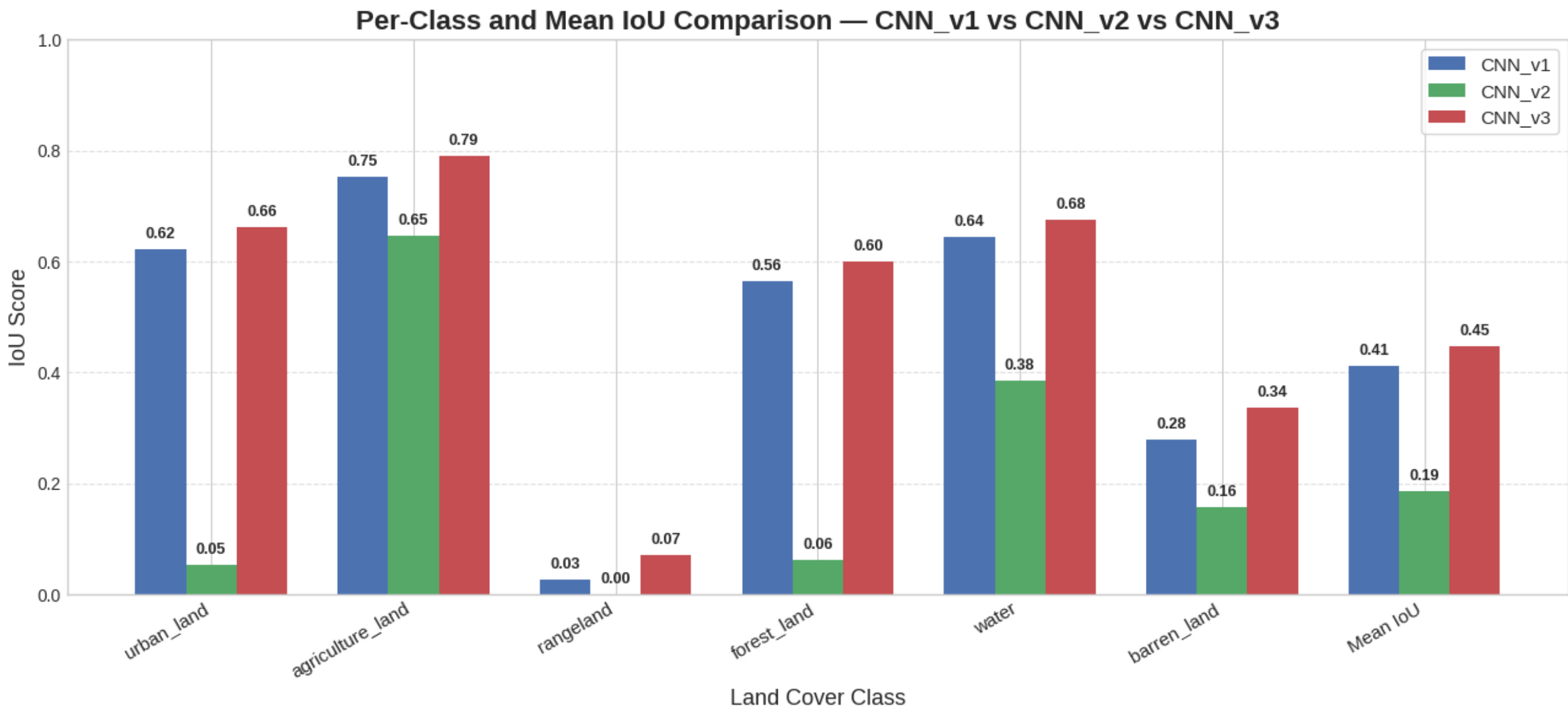

**Fig. 4** Per-class and mean Intersection over Union (IoU) scores for CNN_v1, CNN_v2, and CNN_v3 across the seven DeepGlobe land-cover classes. The horizontal dashed line indicates the mean IoU for each variant.

CNN_v2 performed unpredictably and saw its validation accuracy drop, whereas CNN_v1 and CNN_v3 both showed steady gains in training and validation accuracy across epochs (Fig. 2). For CNN_v3, the close match between training and validation curves points to less overfitting and stronger generalisation. Architectural tweaks particularly the use of SpatialDropout2D and an adaptive learning-rate schedule, helped keep hyperparameter tuning in check, which in turn supported more stable convergence and more reliable predictions (Tarafdar et al., 2025).

CNN_v3 also delivered more consistent classification of urban, agricultural, and forest classes than its predecessors, reflected in the higher diagonal values in the confusion matrices shown in Fig. 3. CNN_v2, by contrast, mixed up rangeland and barren land more frequently an issue tied to its weaker feature representation while CNN_v1 sat in the middle with somewhat lower overall accuracy. Pushpalatha et al., (2025) have argued that fine-tuning network depth and filter size can sharpen spatial pattern discrimination, and they attribute much of the remaining misclassification to class imbalance, specifically the overrepresentation of certain categories.

Fig. 4 shows that CNN_v3 had the highest IoU for agricultural (0.79), urban (0.66), and water (0.68) land-covers. CNN_v2 struggled with rangeland and woodland minority classes due to limited generalisability. Akeboshi et al. (2022) and Marzougui et al. (2025) state that multi-scale convolutional architectures with regularisation improve per-class segmentation. CNN_v3, with higher mean IoU and stable convergence, showed that optimized architecture, adaptive learning rates, and LULC feature regularisation improve segmentation accuracy (Tarafdar et al., 2025; Pushpalatha, 2025).

## AlexNet Results

Table 5 summarises the performance of the AlexNet variants. AlexNet_v2 achieved the highest overall accuracy (0.7649) and mean IoU (0.4064). AlexNet_v1 recorded an accuracy of 0.7335 and mean IoU of 0.3436. AlexNet_v3 attained an accuracy of 0.7493 and a mean IoU of 0.3755.

**Table 5** Performance metrics for AlexNet variants

| **Model** | **Overall Accuracy** | **Mean IoU** | **Precision (macro)** | **Recall (macro)** | **F1-score (macro)** |
|---|---|---|---|---|---|
| **AlexNet_v1** | 0.7335 | 0.3436 | 0.5142 | 0.4322 | 0.4535 |
| **AlexNet_v2** | 0.7649 | 0.4064 | 0.5824 | 0.4937 | 0.5183 |
| **AlexNet_v3** | 0.7493 | 0.3755 | 0.5384 | 0.4622 | 0.4859 |

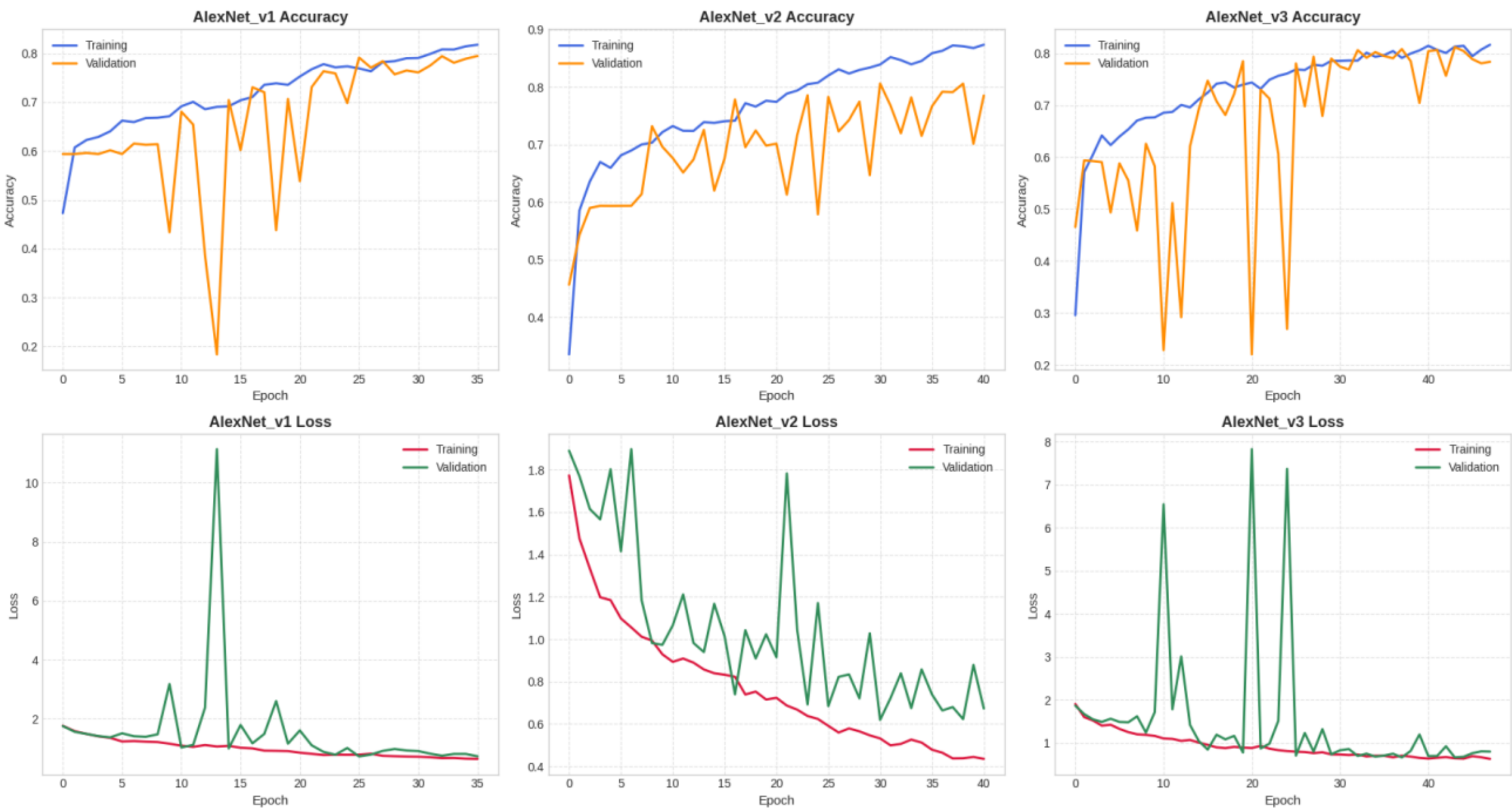

**Fig. 5** Epoch-wise training and validation accuracy and loss trajectories for AlexNet_v1, AlexNet_v2, and AlexNet_v3.

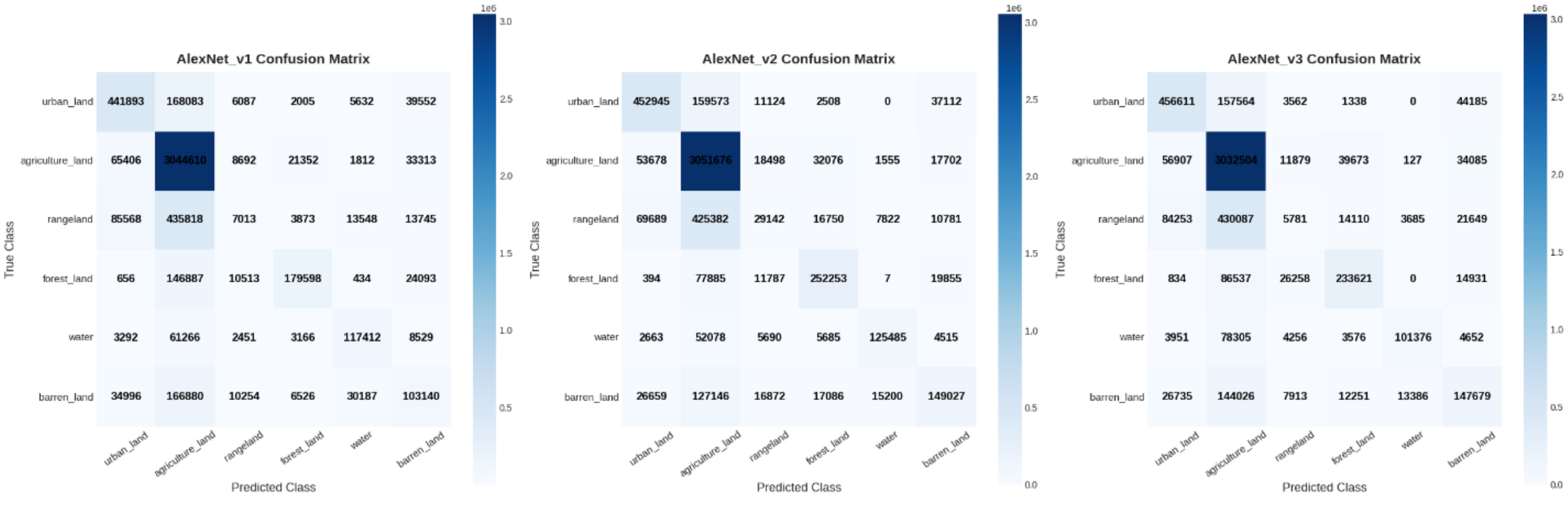

**Fig. 6** Confusion matrices for the three AlexNet variants, illustrating the distribution of predicted labels against ground-truth annotations for each land-cover category on the held-out test set.

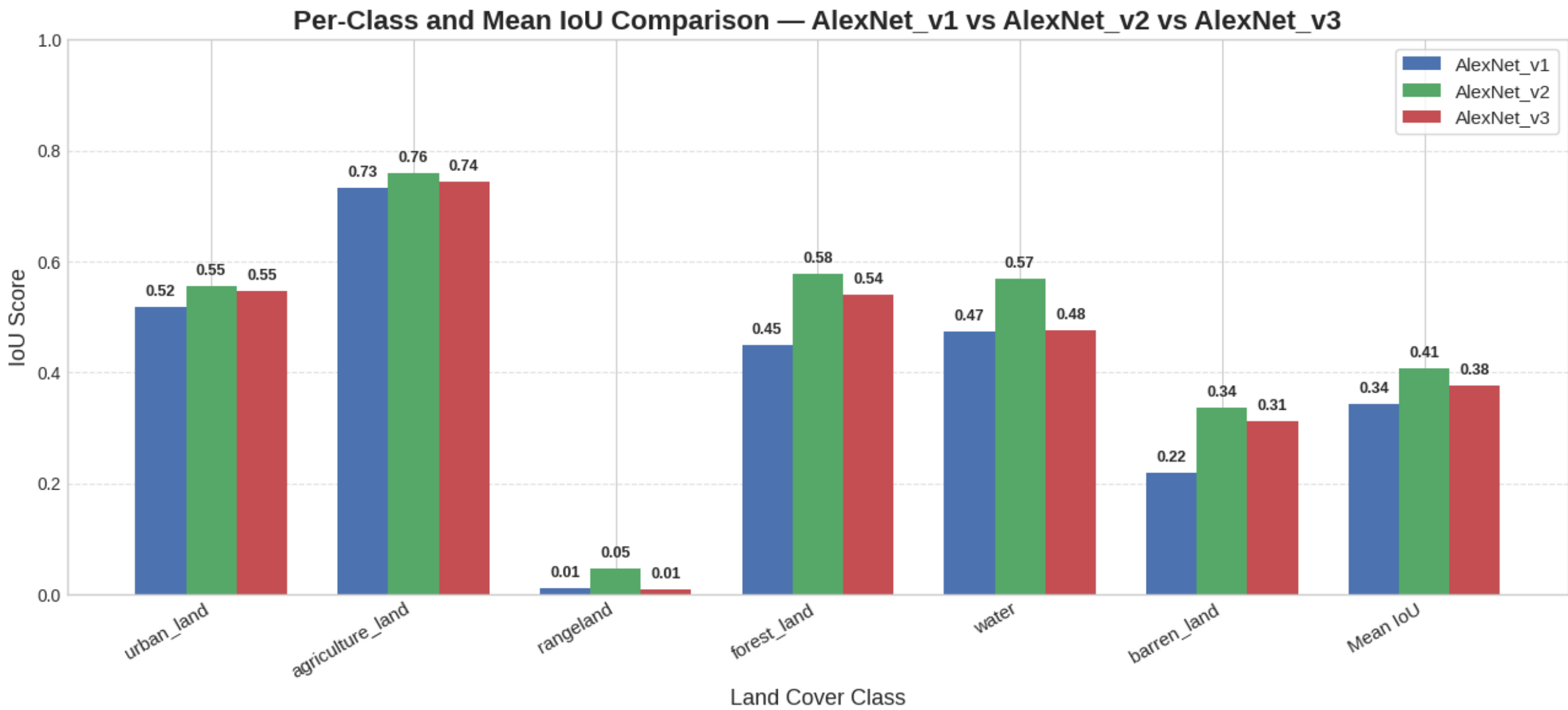

**Fig. 7** Bar chart of per-class IoU and mean IoU values for AlexNet_v1, AlexNet_v2, and AlexNet_v3. Colours correspond to individual land-cover classes, with the final bar in each group representing the mean IoU

With continuous accuracy curves and aligned training and validation behaviors, AlexNet v2 converged more reliably and smoothly (Fig. 5). In AlexNet v1, validation spikes occurred occasionally, but v3's improved regularisation made loss patterns more unstable. Vinaykumar, Babu, and Frnda (2023) showed that AlexNet remains viable for satellite imagery provided the parameters are tuned carefully and regularisation is kept in balance with model complexity.

Fig. 6 presents the confusion matrices. AlexNet_v2 improved diagonal alignment for agriculture, urban, and forest classes compared with v1 and v3. Off-diagonal confusion was most pronounced between rangeland and barren land across all three variants.

AlexNet_v2 had higher IoU values per class for most major categories except water (0.57), forests (0.58), and agricultural (0.76), as shown in Fig. 7. Even though the mean IoU improved slightly from baseline, the three configurations reached a similar representational limit with negligible differences. Reddy and Vimal (2024) found that AlexNet can discriminate well with fine-tuning and regularisation.

# InceptionV3 Results

Table 6 presents the InceptionV3 results. InceptionV3_v2 achieved the highest accuracy (0.7610) and mean IoU (0.4118) among the three variants. InceptionV3_v1 recorded the lowest metrics (accuracy 0.6935, mean IoU 0.2964). InceptionV3_v3 attained an accuracy of 0.7540 and a mean IoU of 0.3907.

**Table 6** Comparative performance metrics across InceptionV3 model variants

| Model | Overall Accuracy | Mean IoU | Precision (macro) | Recall (macro) | F1-score (macro) |
|---|---|---|---|---|---|
| **InceptionV3_v1** | 0.6935 | 0.2964 | 0.4766 | 0.3819 | 0.4108 |
| **InceptionV3_v2** | 0.761 | 0.4118 | 0.5885 | 0.4996 | 0.532 |
| **InceptionV3_v3** | 0.754 | 0.3907 | 0.5629 | 0.4833 | 0.5111 |

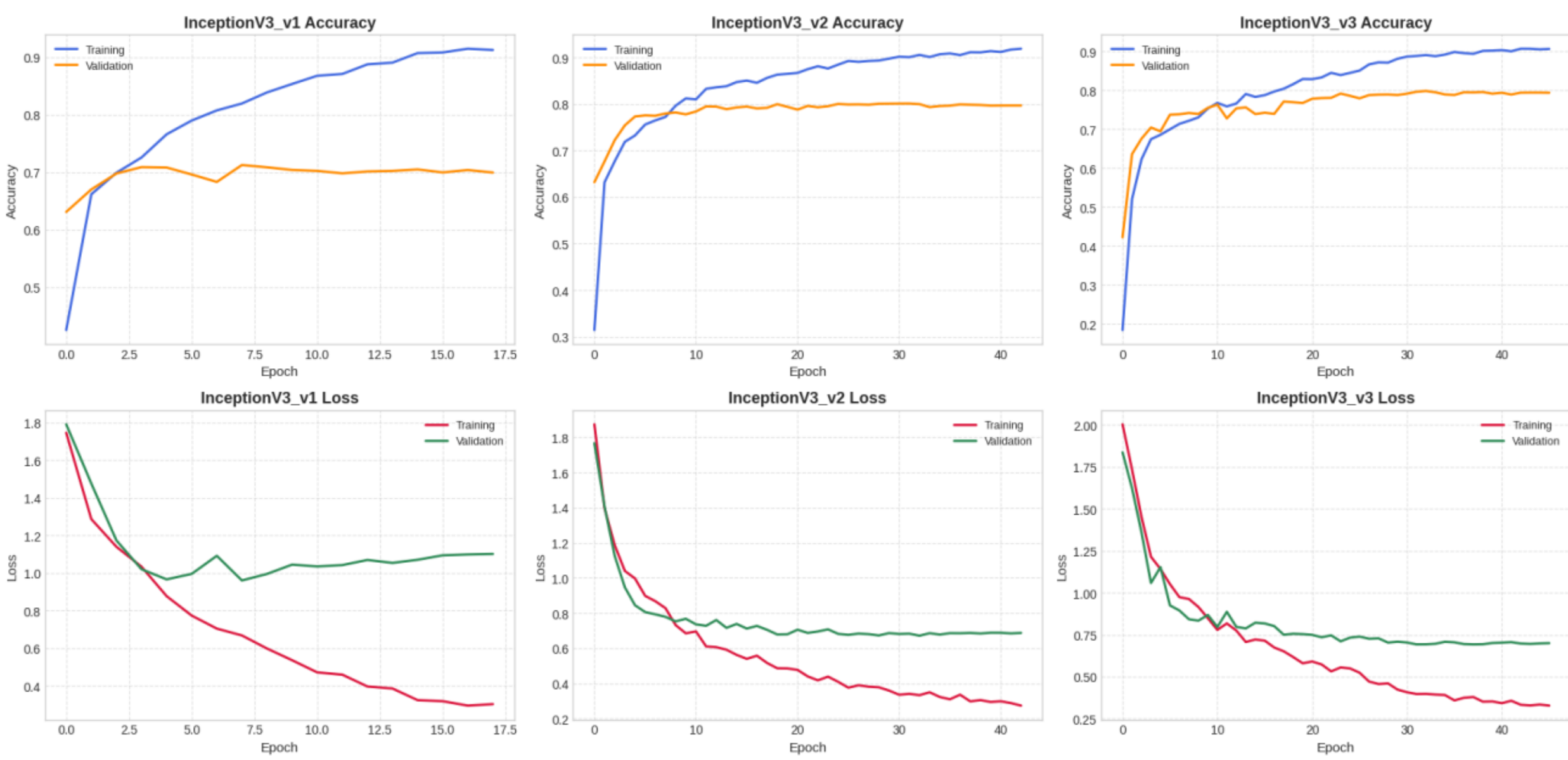

**Fig. 8** Convergence profiles of InceptionV3_v1, InceptionV3_v2, and InceptionV3_v3, showing training and validation accuracy and loss as a function of epoch. Solid and dashed lines distinguish training from validation performance

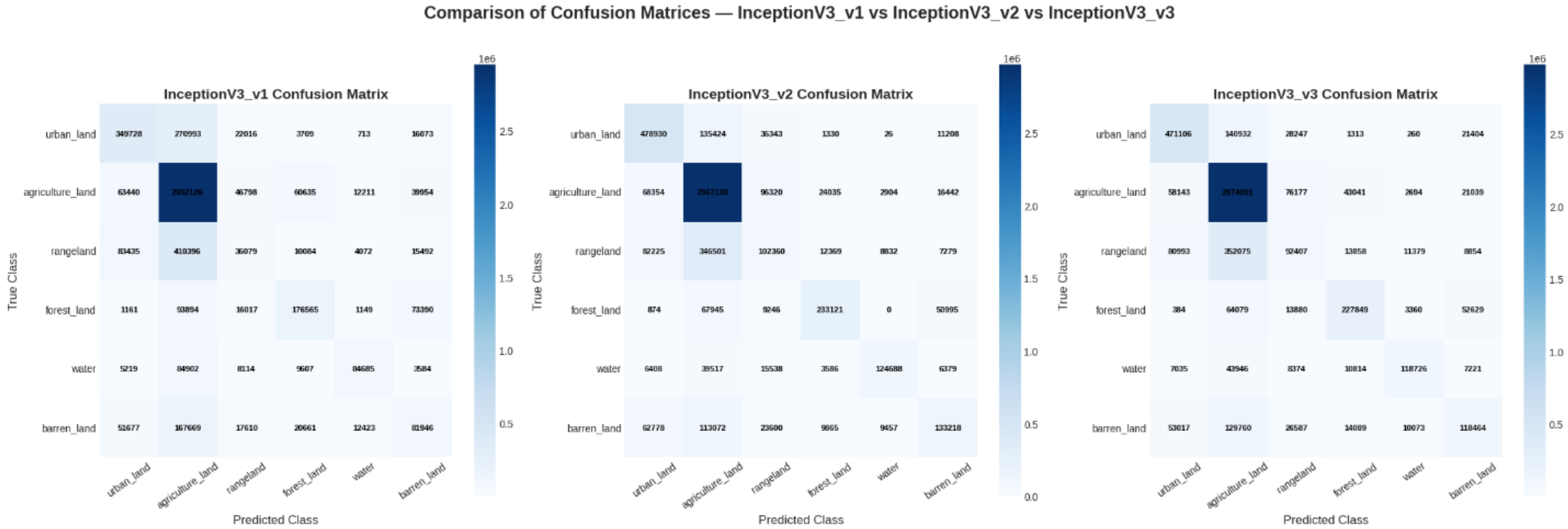

**Fig. 9** Test-set confusion matrices for InceptionV3_v1, InceptionV3_v2, and InceptionV3_v3. Each matrix displays the frequency of predicted class labels relative to the ground-truth reference

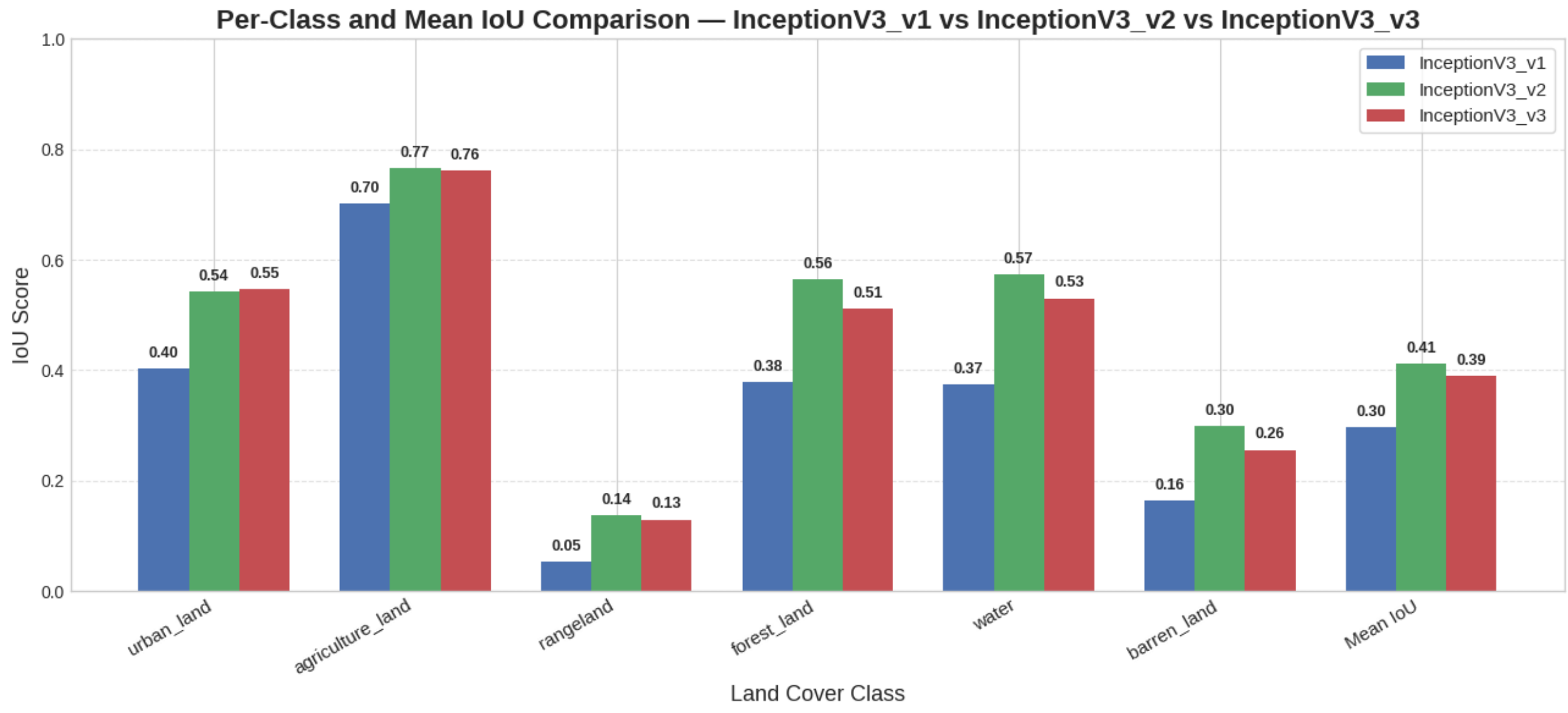

**Fig. 10** Per-class and mean IoU comparison for the three InceptionV3 configurations. Individual bars denote IoU for urban, agricultural, rangeland, forest, water, barren, and unknown classes, with the rightmost bar in each group indicating the mean

InceptionV3_v2 and v3 have sustained convergence with aligned training and validation curves across epochs (Fig. 8). The validation curve for InceptionV3_v1 flattened out early, a pattern that points to frozen weights constraining the learning process. Marzougui et al. (2025) note that adaptive fine-tuning at deeper Inception modules helps mitigate underfitting, stabilise convergence, and enhance differentiation between classes.

Fig. 9's confusion matrices show InceptionV3_v2 improved urban, forest, and agriculture predictions and reduced barren and rangeland misclassifications. The model represents the spatial-spectral hierarchy well. As expected given

its limited feature learning, InceptionV3_v1 confused land-type boundaries more than InceptionV3_v3, which had slightly better accuracy but worse recall. According to Alem and Kumar (2022), deep inception networks sharpen discrimination among visually overlapping land classes when regularisation is tuned appropriately.

Fig. 10 shows that InceptionV3_v2 was the most stable solution across all classes, with intersection-over-union values of 0.77 for agricultural, 0.56 for forest, and 0.57 for water. Fine-tuned layers have adapted to DeepGlobe's spatial variety, as IoU outperforms InceptionV3_v1. Fayaz et al. (2024) showed that Inception-based architectures outperform CNNs in contextual complexity and multi-scale pattern extraction.

# MobileNetV2 Results

Table 7 reports on MobileNetV2 performance. MobileNetV2_v1 achieved the highest overall accuracy (0.7906) and mean IoU (0.4625) across the entire study. MobileNetV2_v2 recorded an accuracy of 0.7666 and mean IoU of 0.4089. MobileNetV2_v3 attained an accuracy of 0.7589 and a mean IoU of 0.4069.

**Table 7** Performance metrics for MobileNetV2 variants

| Model | Overall Accuracy | Mean IoU | Precision (macro) | Recall (macro) | F1-score (macro) |
|---|---|---|---|---|---|
| **MobileNetV2_v1** | 0.7906 | 0.4625 | 0.6069 | 0.5641 | 0.5786 |
| **MobileNetV2_v2** | 0.7666 | 0.4089 | 0.5958 | 0.4918 | 0.5197 |
| **MobileNetV2_v3** | 0.7589 | 0.4069 | 0.5815 | 0.491 | 0.5253 |

Training and Validation Accuracy & Loss — MobileNetV2_v1 vs v2 vs v3

MobileNetV2_v1 — Accuracy

MobileNetV2_v2 — Accuracy

MobileNetV2_v3 — Accuracy

MobileNetV2_v1 — Loss

MobileNetV2_v2 — Loss

MobileNetV2_v3 — Loss

**Fig. 11** Accuracy and loss curves during training for MobileNetV2_v1, MobileNetV2_v2, and MobileNetV2_v3. Training trajectories are plotted as solid lines and validation trajectories as dashed lines over the course of optimisation

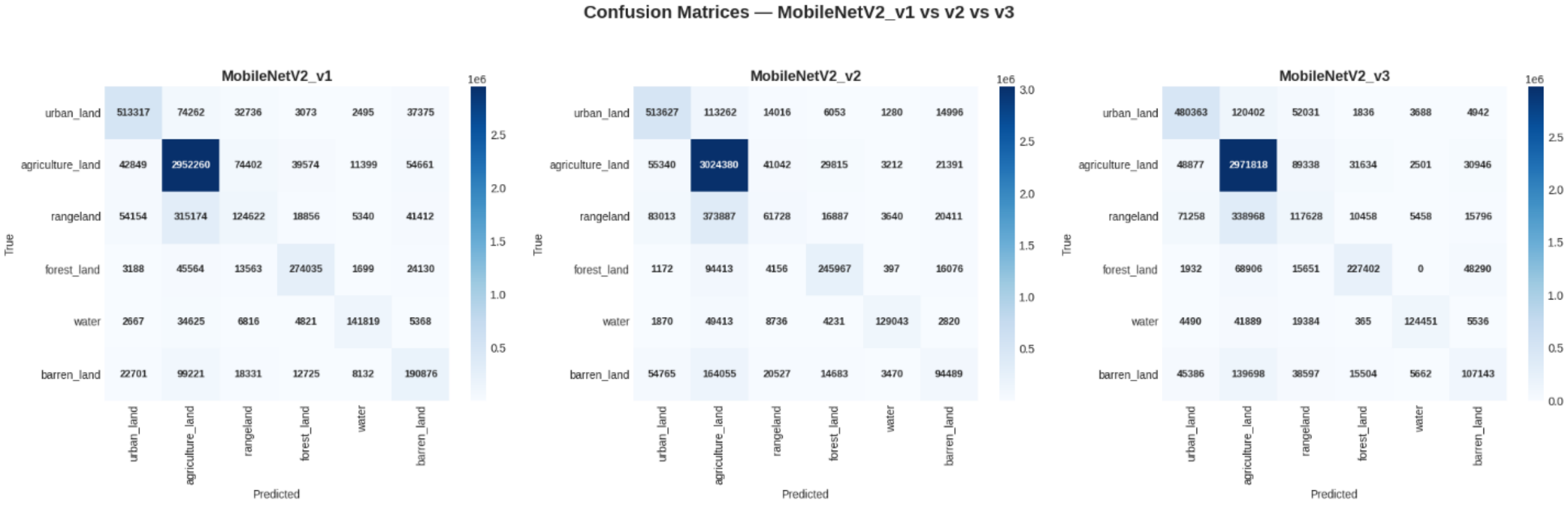

**Fig. 12** Confusion matrices comparing predicted and true class distributions for MobileNetV2_v1, MobileNetV2_v2, and MobileNetV2_v3 on the DeepGlobe test imagery

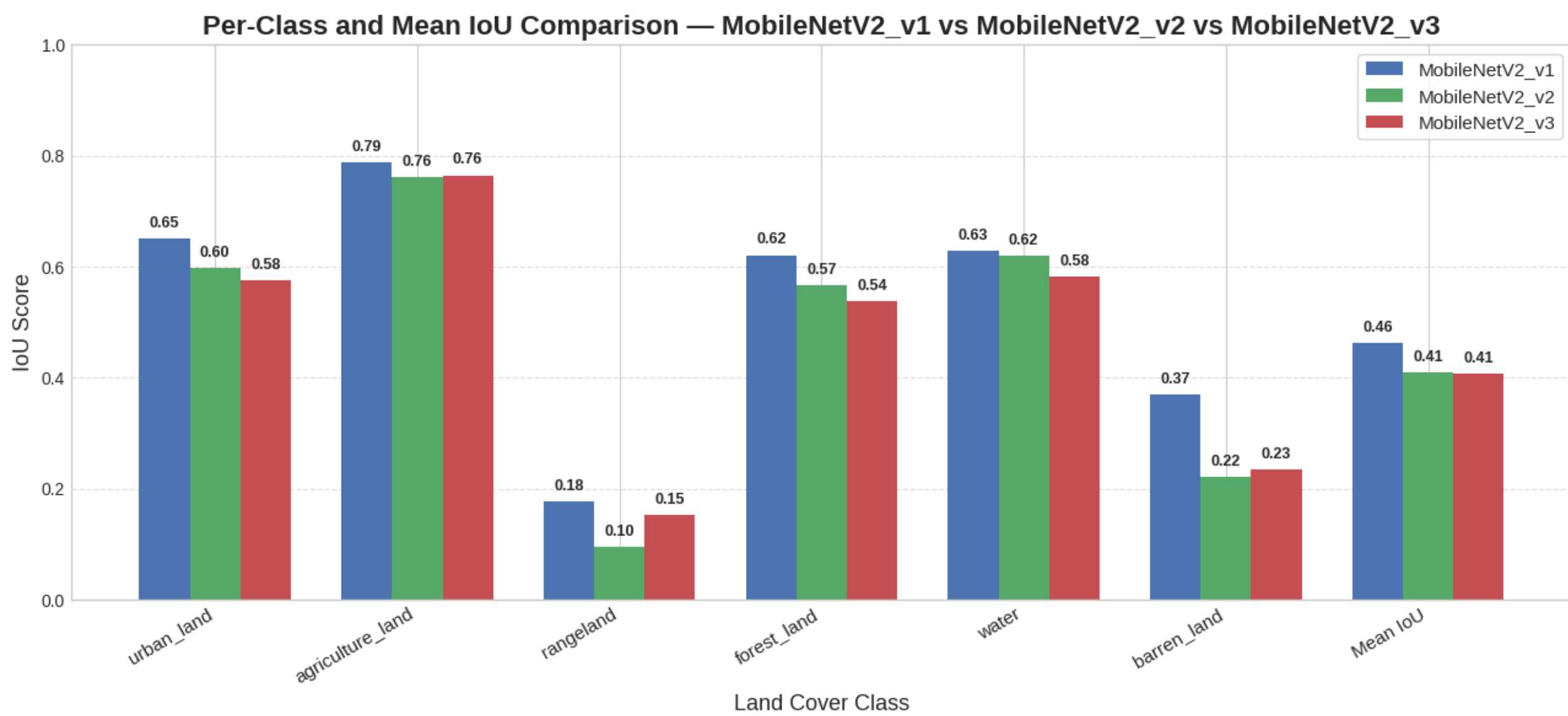

**Fig. 13** Per-class IoU and mean IoU values for the three MobileNetV2 variants. The grouped bars show segmentation accuracy for each land-cover category, with the mean IoU highlighted in the final column of each group

Fig. 11 shows smooth convergence with little overfitting for all three layouts. Regularisation and gradient propagation kept MobileNetV2_v1 accurate during training and validation. MobileNetV2_v2 came closer but plateaued earlier than MobileNetV2_v3, which had validation loss oscillations due to dropout. Thapliyal et al. (2024) say depthwise separable convolutions and adaptive layer scaling achieve training stability and performance across MobileNet versions.

Fig. 12 shows confusion matrices showing classification accuracy across important land-cover categories, especially urban, forest, and agricultural. MobileNetV2_v1 distinguished urban and barren better than v2. According to Anjali et al. (2024), lightweight CNNs may have trouble distinguishing spectrally similar areas without attention mechanisms. Most misclassifications occurred between visually identical textures like rangeland and barren land.

Fig. 13 indicates that MobileNetV2_v1 achieved a higher IoU across most land-cover categories, though it fell short for water (0.63), forests (0.62), and agricultural land (0.79). The narrow gap in class-wise IoU between MobileNetV2_v2 and MobileNetV2_v3 points to consistent performance and reliability across these architecture variants. Barman and Susan (2024) have shown that MobileNetV2, alongside other lightweight transfer learning models, can still deliver accurate predictions even when computational resources are limited.

# VGG16 Results

Table 8 summarises the VGG16 results. VGG16_v2 achieved the highest mean IoU (0.4244) and F1-score (0.5411). VGG16_v3 attained the highest overall accuracy (0.7690) and precision (0.6319). VGG16_v1 recorded the lowest metrics (accuracy 0.7274, mean IoU 0.3512).

**Table 8** Performance metrics for VGG16 variants

| Model | Overall Accuracy | Mean IoU | Precision (macro) | Recall (macro) | F1-score (macro) |
|---|---|---|---|---|---|
| **VGG16_v1** | 0.7274 | 0.3512 | 0.5282 | 0.431 | 0.4598 |
| **VGG16_v2** | 0.7653 | 0.4244 | 0.6357 | 0.4996 | 0.5411 |
| **VGG16_v3** | 0.769 | 0.4214 | 0.6319 | 0.4887 | 0.5374 |

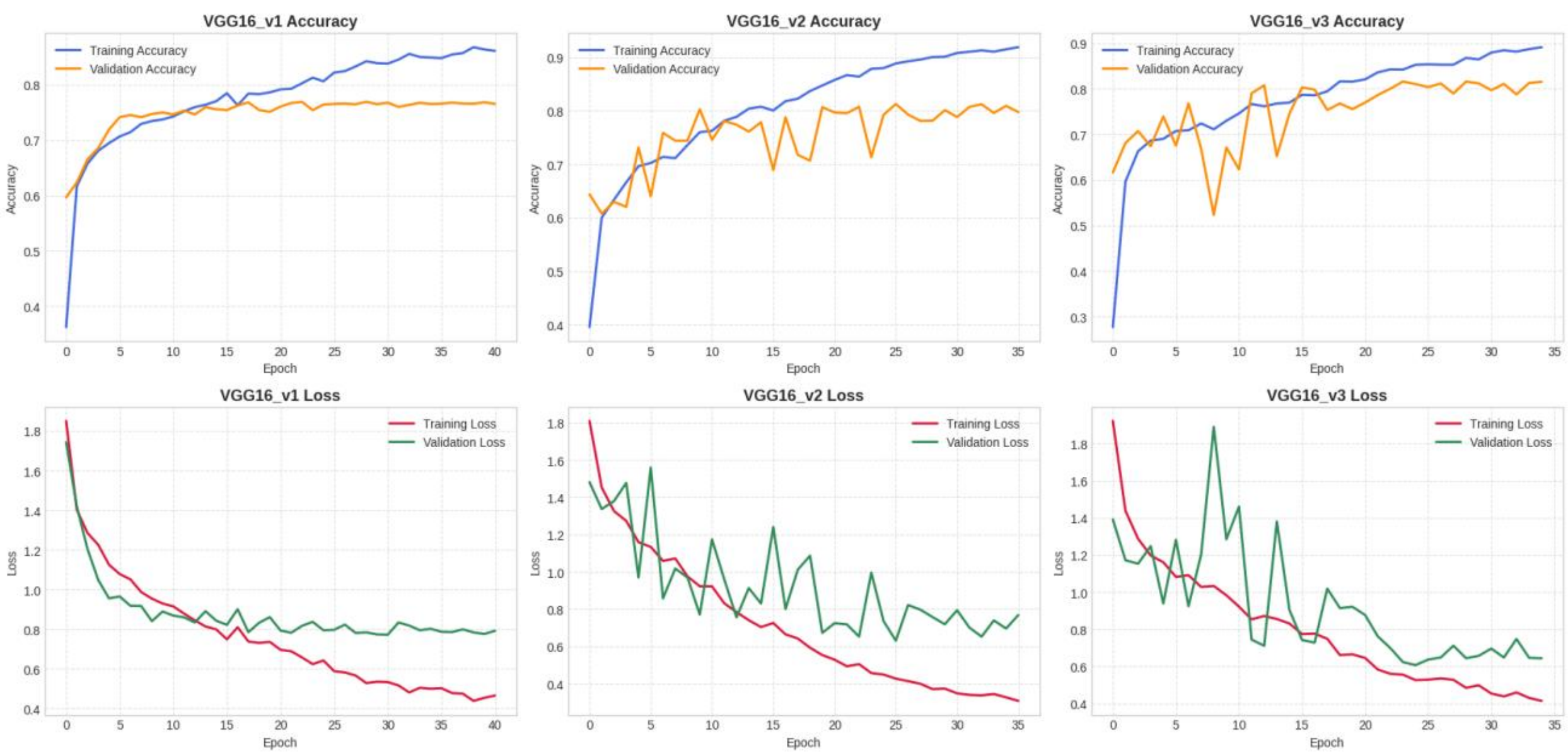

**Fig.14** Training and validation accuracy and loss evolution for VGG16_v1, VGG16_v2, and VGG16_v3. Solid lines correspond to training data and dashed lines to validation data

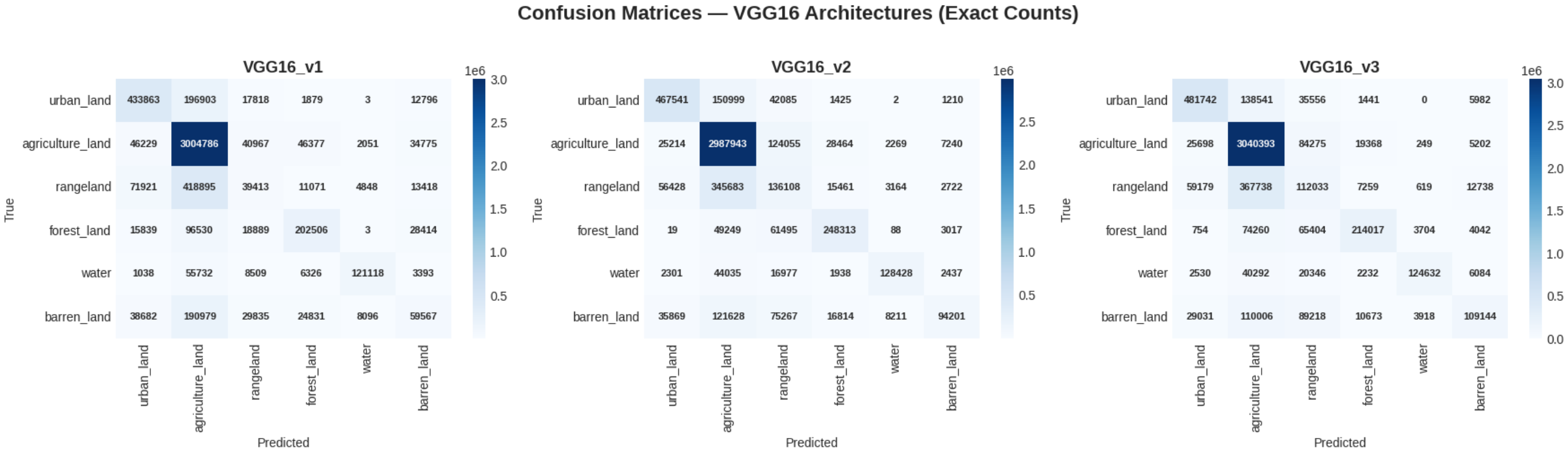

**Fig. 15** Confusion matrices for VGG16_v1, VGG16_v2, and VGG16_v3, depicting the alignment between model predictions and ground-truth labels across all seven land-cover classes

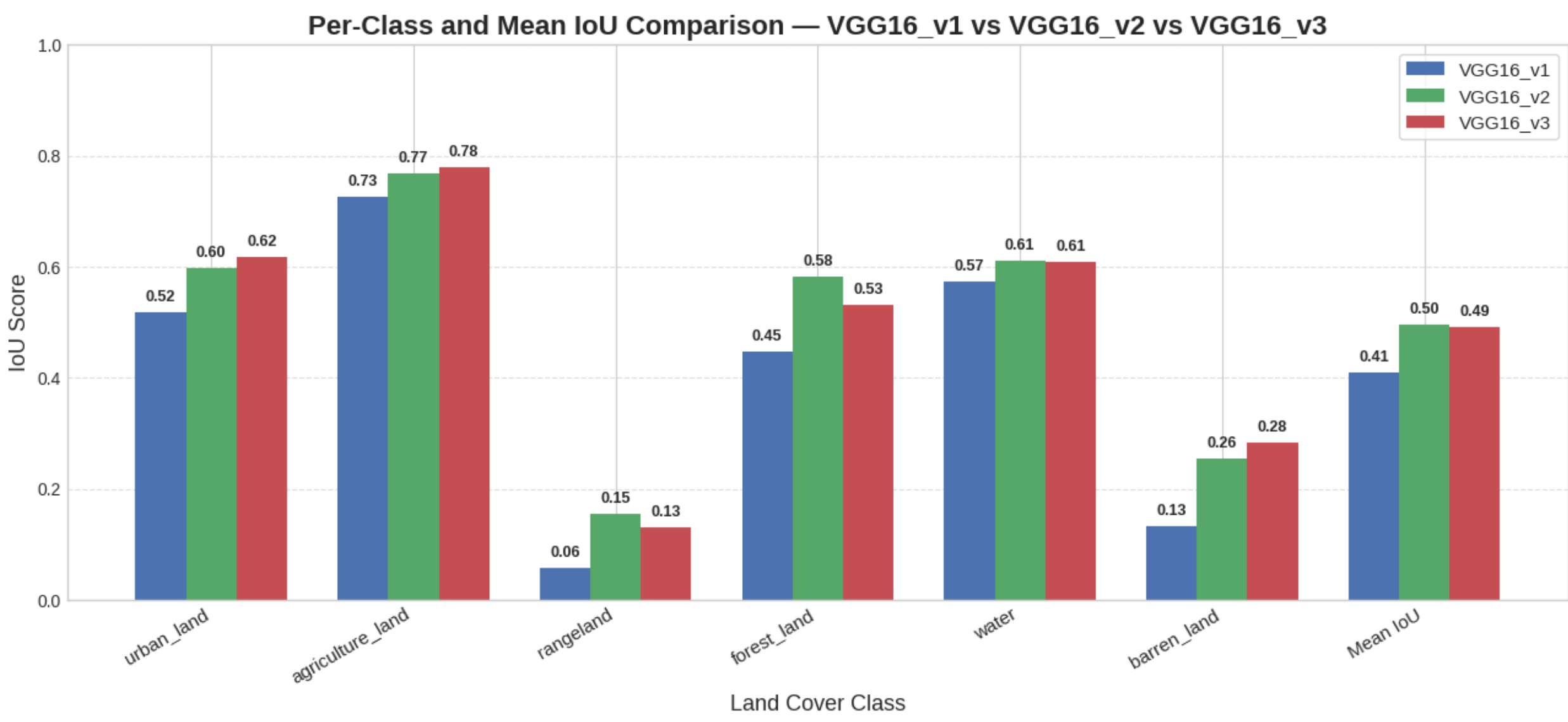

**Fig. 16** Per-class IoU scores and mean IoU for the three VGG16 configurations. Each cluster of bars represents one variant, with individual land-cover classes color-coded and the mean IoU shown as the final bar

Fig. 14 shows that all three configurations converged steadily across training epochs, with VGG16_v2 and VGG16_v3 pulling ahead on validation accuracy. The optimisation appeared to work well: the gap between training and validation losses stayed narrow, and the loss curves traced a smooth, consistent downward path. Kumar et al. (2024) argue that VGG-based encoders preserve hierarchical spatial representations, which help anchor feature convergence when working with high-dimensional remote-sensing data.

The confusion matrices in Fig. 15 show that forest, urban, and farmland types are accurately classified. VGG16_v1 confused rangeland and urban land, but VGG16_v2 better classified water and barren areas. Adaptive feature extraction and refined weight modification improve classification accuracy in fine-tuned VGG16 architectures (Ahangarha et al., 2024).

Most terrain types improve per-class IoU as architecture versions improve (Fig. 16). In the agricultural (0.78) and forest (0.53) datasets, VGG16_v3 segmented well with balanced IoU across all classes. (Dey et al., 2024) The model's hierarchical convolutional layers and pre-trained filters are praised, and the model's land type versatility supports this.

# Model Deployment and Predictive Demonstration

Final MobileNetV2_v1 was deployed on unseen DeepGlobe test imagery to test real-world generalization. The trained segmentation network created pixel-wise class prediction masks from 256x256 satellite images after normalization. Per-class pixel counts and the dominant land-cover category were quantified and identified for operational land management workflows.

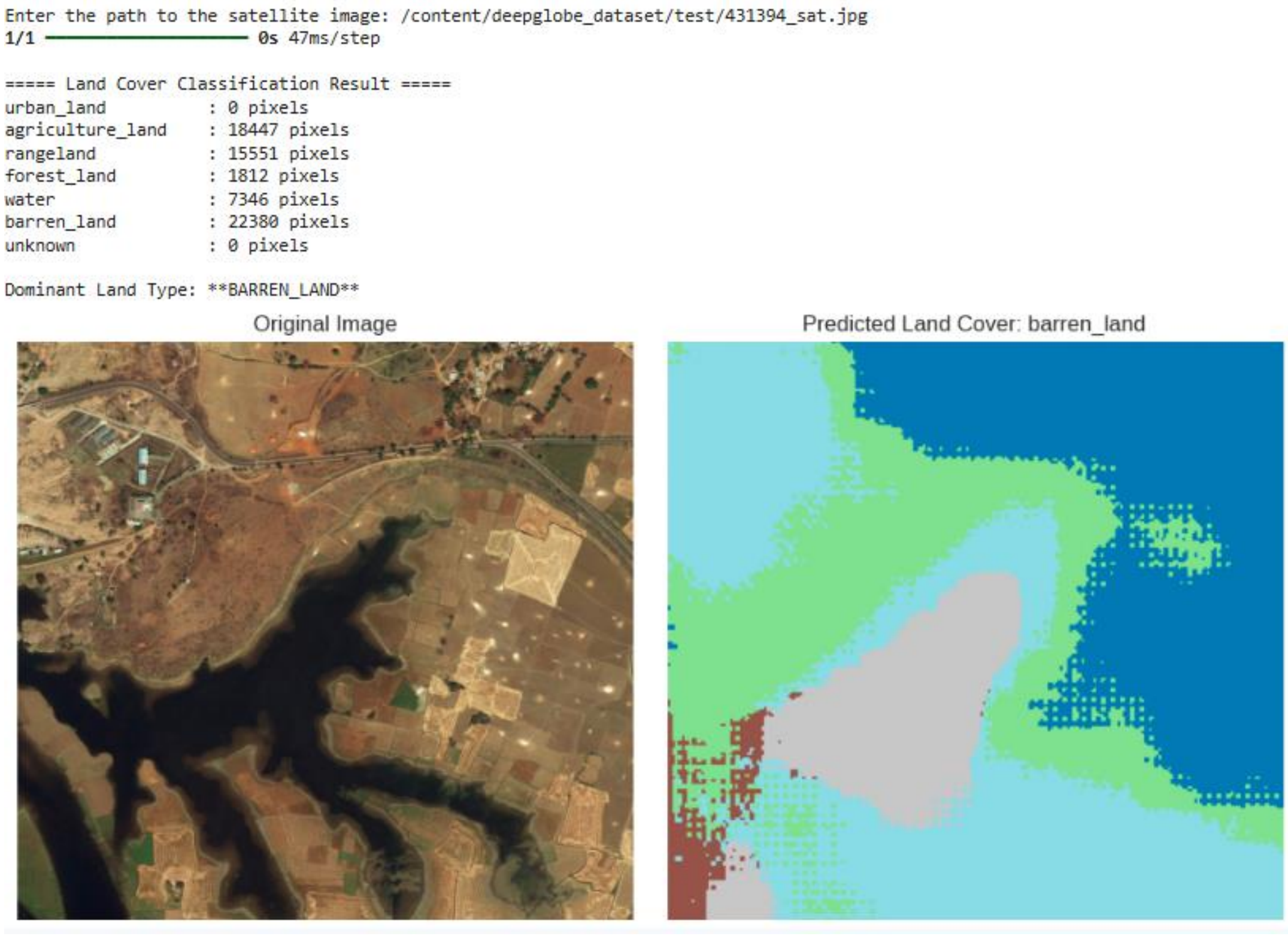

**Fig. 17** Pixel-wise land-cover segmentation output from MobileNetV2_v1 on an unseen DeepGlobe satellite image. The input RGB scene (left) contains barren and agricultural terrain adjacent to a water body; the corresponding predicted mask (right) assigns each pixel to one of seven land-cover categories according to the class colour legend

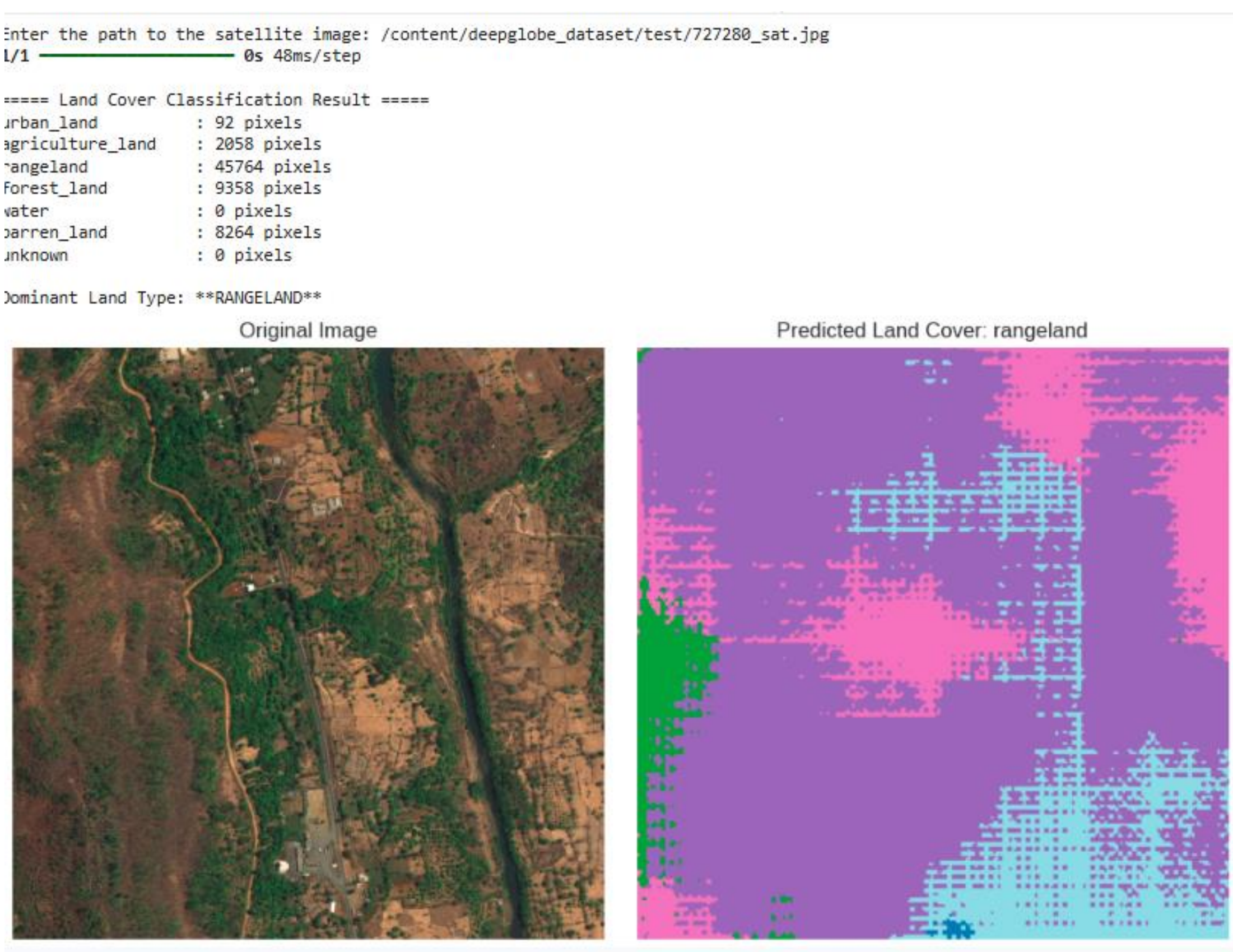

**Fig. 18** Pixel-wise land-cover segmentation output from MobileNetV2_v1 on a second unseen DeepGlobe satellite image. The input RGB scene (left) is dominated by rangeland with interspersed forest patches; the corresponding predicted mask (right) shows the spatial distribution of predicted classes across the scene

Fig.s 17 and 18 illustrate the model's deployment performance on two held-out test images. In Fig. 17, the model accurately segments expansive barren and agricultural regions bordering a water body, with crisp boundary delineation between land and water. Fig. 18 reveals strong spatial coherence in rangeland classification, with forest patches cleanly separated despite spectral similarity between the two classes. The minimal class overlap and consistent color-coded segmentation across both scenes confirm robust generalization beyond the training distribution, aligning with the quantitative mean IoU of 0.4625. These results demonstrate that MobileNetV2_v1 achieves an effective accuracy–efficiency trade-off, positioning it as a viable solution for scalable, resource-constrained land-cover mapping applications.

# Conclusion

This benchmark compares five architectures CNN, AlexNet, InceptionV3, MobileNetV2, and VGG16 for semantic segmentation on the DeepGlobe dataset, with each evaluated through three progressively optimized iterations under identical conditions. MobileNetV2_v1 achieved the optimal efficiency–accuracy frontier, attaining the highest overall accuracy (0.7906) and mean IoU (0.4625) at merely 24.98 MB, outperforming substantially deeper alternatives including InceptionV3_v2 (125.17 MB, accuracy 0.7610) and VGG16_v2 (71.13 MB, accuracy 0.7653). Three key insights emerge: first, MobileNetV2's depthwise separable convolutions enable efficient multi-scale feature extraction suited to resource-constrained deployment; second, aggressive regularisation yields diminishing returns, with InceptionV3_v3 and VGG16_v3 exhibiting marginal or negative gains relative to partially fine-tuned variants, indicating overfitting risk in deeper networks on moderate-sized datasets; third, persistent rangeland–barren confusion across all architectures reflects spectral similarity and minority-class underrepresentation unresolvable through architectural choice alone. Deployment on held-out imagery confirms robust generalization, with spatially coherent masks validating operational applicability for agricultural monitoring, urban expansion tracking, and water resource assessment.

There are some limitations. First, deliberate exclusion of data augmentation and class-imbalance correction, while ensuring controlled comparison, caps achievable accuracy and may understate architectures benefiting from augmented training. Second, the 256 × 256 resolution sacrifices sub-meter detail critical for fine-grained delineation, potentially contributing to persistent rangeland–barren confusion via texture loss. Third, ImageNet pretraining introduces a domain gap between natural-scene and overhead satellite feature spaces, limiting transfer efficacy for spectrally distinctive classes. Fourth, single-dataset evaluation on DeepGlobe constrains geographic generalisability. Finally, computational benchmarking on cloud GPUs leaves edge-deployment latency and energy consumption unquantified. Future work should integrate attention mechanisms, multispectral inputs, and embedded hardware validation to address these boundaries.